\documentclass{article}

\usepackage{PRIMEarxiv}

\usepackage[utf8]{inputenc} 
\usepackage[T1]{fontenc}    
\usepackage{hyperref}       
\usepackage{url}            
\usepackage{booktabs}       
\usepackage{amsfonts}       
\usepackage{nicefrac}       
\usepackage{microtype}      
\usepackage{lipsum}
\usepackage{fancyhdr}       
\usepackage{graphicx}       
\graphicspath{{media/}}     
\usepackage{amsmath}

\title{Fast Posterior Estimation of Cardiac Electrophysiological Model Parameters via Bayesian Active Learning
}

\author{
  Md Shakil Zaman\\
  Rochester Institute of Technology\\
  Rochester, NY, USA\\
  \texttt{mz1482@rit.edu} \\
      \And
  Jwala Dhamala \\
  Rochester Institute of Technology\\
  Rochester, NY, USA\\
  \texttt{jd1336@rit.edu} \\

  \And
  Pradeep Bajracharya \\
  Rochester Institute of Technology\\
  Rochester, NY, USA\\
  \texttt{pb8294@rit.edu} \\
  
  \And
  John L. Sapp \\
  Dalhousie University\\
  Halifax, Canada\\
  \texttt{john.sapp@nshealth.ca} \\
    \And
   B.~Milan~Hor{\'a}cek \\
  Dalhousie University\\
  Halifax, Canada\\
  \texttt{milan.horacek@dal.ca} \\
  
    \And
  Katherine~C.~Wu \\
  Johns Hopkins University\\
   Baltimore, MD, USA\\
  \texttt{kwu@jhmi.edu} \\
    \And
   Natalia~A.~Trayanova \\
  Johns Hopkins University\\
   Baltimore, MD, USA\\
  \texttt{ntrayanova@jhu.edu} \\

    \And
   Linwei Wang\\
  Rochester Institute of Technology\\
  Rochester, NY, USA\\
  \texttt{linwei.wang@rit.edu} \\
}

\begin{document}
\maketitle

\begin{abstract}
Probabilistic estimation of cardiac electrophysiological model parameters serves an important step towards model personalization and uncertain quantification. The expensive computation associated with these model simulations, however, makes direct Markov Chain Monte Carlo (MCMC) sampling of the posterior probability density function (pdf) of model parameters computationally intensive. Approximated posterior pdfs resulting from replacing the simulation model with a computationally efficient surrogate, on the other hand, have seen limited accuracy. In this paper, we present a Bayesian active learning method to directly approximate the posterior pdf function of cardiac model parameters, in which we intelligently select training points to query the simulation model in order to learn the posterior pdf using a small number of samples. We integrate a generative model into Bayesian active learning to allow approximating posterior pdf of high-dimensional model parameters at the resolution of the cardiac mesh. We further introduce new acquisition functions to focus the selection of training points on better approximating the shape rather than the modes of the posterior pdf of interest. We evaluated the presented method in estimating tissue excitability in a 3D cardiac electrophysiological model in a range of synthetic and real-data experiments. We demonstrated its improved accuracy in approximating the posterior pdf compared to Bayesian active learning using regular acquisition functions, and substantially reduced computational cost in comparison to existing standard or accelerated MCMC sampling.
\end{abstract}


\section{Introduction}
With advanced technologies in medical imaging and image analysis, computational models can now closely replicate the physiology of a human heart \cite{taylor2009patient, morris2016computational}. As these models are virtual in nature, they have the potential to enable prediction, diagnosis, and treatment planning of certain conditions of a patient's heart with little to no harm to the patient \cite{prakosa2018personalized, arevalo2016arrhythmia, sermesant2012patient, cronin20192019, zahid2016feasibility}. However, while the geometry of a specific patient heart can be depicted with increasing accuracy, patient-specific physiology remains a challenge. A main difficulty arises from the need to customize patient-specific material properties \cite{taylor2009patient, neal2010current}, which are typically spatially varying throughout the 3D organ and may change over time for the same individual. At the same time, they often cannot be directly measured in high resolution, but have to be estimated from relatively limited measurements. This results in a challenging inverse problem for estimating high dimensional (HD) unknown parameters of a complex, nonlinear, and computationally-expensive forward model that relates the unknown parameters to measurements.

There are two general approaches to this inverse problem: deterministic optimization and probabilistic inference. In deterministic optimization, we seek a single optimal value of the unknown model parameter that will minimize the mismatch between the model output and the measurement data \cite{wong2012strain, sermesant2012patient, wong2015velocity, balaban2018vivo, mineroff2019optimization, yang2015estimation, barone2020efficient, barone2020experimental, balaban2018vivo}. These estimates however do not take into account the uncertainty in the measurement data, nor can they offer insights into the presence of non-unique solutions that can match the same data. These can be overcome by probabilistic inference of the posterior pdf of the model parameters given available measurements.

Existing approaches to the probabilistic estimation of model parameters are generally based on Markov Chain Monte Carlo (MCMC) sampling. The computation expense of the forward simulations of these models, however, makes MCMC infeasible due to the reliance on a large number of sampling each requiring a simulation run. Approaches to accelerating such sampling can be loosely divided into two categories. On one hand, a variety of hybrid sampling methods has been developed, which accelerates random sampling using information about the target pdf such as its gradient \cite{roberts1996exponential, neal2010mcmc} and Hessian matrix \cite{martin2012stochastic}. These information however are often difficult to extract from the posterior pdf involving a complex simulation model. On the other hand, it is possible to construct a computationally-efficient approximation, \textit{i.e.}, surrogate model, of the expensive simulation process, such that the related pdfs become substantially faster to sample. These surrogate models may be physics-based reduced-order modeling \cite{lassila2013reduced}, or data-driven approximations such as Gaussian process (GP) \cite{kennedy2000predicting, rasmussen2003gaussian} and polynomial chaos \cite{spanos1989stochastic, xiu2003modeling, marzouk2009dimensionality}. Directly sampling the surrogate-based posterior pdf, however, may lead to limited accuracy due to the difficulty to build a globally-accurate approximation of a complex nonlinear simulation model. In our previous work, we attempted to mitigate this issue by using this surrogate-based pdf to accelerate, rather than replacing, the sampling of the actual pdf \cite{dhamala2018quantifying}. Specifically, this was achieved by a two-stage MCMC strategy where the surrogate-based pdf works as a proposal distribution to increase the acceptance rate of sampling \cite{dhamala2018quantifying}. While this ensures the accuracy of posterior sampling, the reduction of the computation becomes limited due to the fundamental reliance on sampling the original pdf involving expensive simulation processes. 

In this paper, we develop a Bayesian active learning approach to provide an accurate surrogate model of the posterior pdf of simulation model parameters such that there is no need of further MCMC sampling of the original computational-intensive pdf. This is achieved with two key innovations. First, unlike most existing approaches that rely on learning a surrogate of the simulation model over the prior distribution of the parameter space \cite{dhamala2018quantifying}, we propose to directly learn a surrogate of the posterior pdf. We formulate this posterior pdf estimation as an active learning problem where we intelligently select a minimal number of training points focused on the posterior support of the parameter space. Second, we present new acquisition functions during the active learning to utilize the shape of the posterior pdf to improve the selection of training points. To enable this active posterior estimation over a high-dimensional parameter space, we further combine it with our previously-developed approach that uses generative modeling of the high-dimensional parameter space  \cite{dhamala2018high} to embed active learning of a high-dimensional posterior pdf into a low-dimensional space.

While our method is generally applicable to posterior estimation of HD parameters in complex models, in this paper it was applied to estimate tissue excitability as parameters of the cardiac electrophysiological model. Experiments were performed on three different groups of data: simulation data with a synthetic setting of abnormal tissues, simulation data generated from a high-fidelity biophysics model blinded to the model used in the posterior estimation, and real data obtained from patients with infarcts derived from \textit{in-vivo} voltage mapping data. In the synthetic group, we compared results with direct MCMC sampling of the original posterior pdf, two-stage MCMC method \cite{dhamala2017spatially}, and direct MCMC sampling of the surrogate pdf learned using regular Bayesian active learning. The results showed that the presented method was able to use $0.6\%$ computation of the direct or two-stage MCMC methods to deliver an accurate estimation of the posterior pdf, with significantly improved accuracy compared to using regular Bayesian active learning. In the other two sets of experiments, we evaluated and interpreted the mean, mode, and uncertainty of the estimated tissue excitability using \textit{in-vivo} magnetic resonance (MR) scar imaging or voltage mapping data.

The key contributions of this work can be summarized as:
\begin{enumerate}
    \item We present a Bayesian active learning approach for fast approximation of the posterior pdf of the parameters of expensive simulation models, with acquisition functions designed to improve the accuracy of the approximation in order to remove the need of subsequent MCMC of the original computationally-expensive pdf.

    \item 
    We leverage our previously developed approach \cite{dhamala2018high} to embed the active learning over high-dimensional space into a low-dimensional manifold, enabling active posterior inference over high-dimensional model parameters representing spatially-varying tissue excitability.

    \item We thoroughly evaluated the performance of the presented method in comparison with existing works in probabilistic parameter estimation in cardiac electrophysiological models, both in synthetic data involving MCMC sampling as reference data, and in real data involving magnetic resonance imaging (MRI) scar imaging and \textit{in-vivo} voltage mapping as reference data.
\end{enumerate}

The rest of the paper is organized as follows. In Section \ref{sec:review}, we review related works in detail and in Section \ref{sec:bck}, we present background of this study. In Section \ref{sec:method}, we present our methodological developments. We present experiments and results for both synthetic and real data from the cardiac electrophysiology system in Section \ref{sec:experiments}. Finally, we give some concluding remarks with limitations and future scope.

\section{Literature Review}
\label{sec:review}
\subsection{Probabilistic Parameter Estimation in Complex Models}

For complex models where the posterior pdf of model parameters are analytically intractable, the area of estimating parameters largely depends on MCMC sampling. Metropolis-Hastings (MH) sampling, Gibbs sampling, and many more classical MCMC methods are developed in \cite{metropolis1949monte, hastings1970monte, geman1984stochastic, gelfand1990sampling, gelfand1992bayesian} and applied in different areas to estimate parameter uncertainty \cite{andrieu2003introduction}. The reason for the extensive use of MCMC is that it can deal with HD parameters, non-linear relation between parameters and observations, and noisy data. However, these properties also make it very slow as, by design, the sampling takes a large number of simulations to converge. With rapid developments of parallel computing, parallel MCMC to accelerate the computation is proposed in \cite{brockwell2006parallel, byrd2010parallel, wang2014parallel} but these can improve neither the convergence rate nor reduce the number of simulations needed. In exploring uncertainty on HD parameters, reversible jump MCMC is used in \cite{brooks1998markov}. Combination of differential evolutions to have subspace exploration is used in \cite{laloy2012high}, while non-differential sparse priors are developed in \cite{cai2018uncertainty}. Gradient and Hessian information of the pdfs are used to accelerate sampling even with poor initial models in \cite{zhao2019gradient}, although these information are nontrivial to extract when the pdf contains complex simulation models.

Alternatively, surrogate models have been widely employed to generate a computational-efficient approximation of the posterior pdf that can be faster to sample. Polynomial chaos \cite{spanos1989stochastic, xiu2003modeling, knio2006uncertainty} and Gaussian process (GP) \cite{kennedy2000predicting, rasmussen2003gaussian} are pioneers in surrogate modelling. In \cite{adams2008gaussian, konukoglu2011efficient, schiavazzi2016uncertainty, gramacy2008bayesian}, to build posterior pdf, GP surrogate is built of the pdf at first, and then MCMC sampling is done from that to avoid expensive simulations. It is however difficult to obtain an approximation of a complex simulation model over the prior parameter space. As a result, when direct sampling of the surrogate pdf is substantially more efficient than sampling the original pdf, the accuracy is often largely compromised \cite{dhamala2018quantifying}. Recently, hybrid approaches are emerging that use the surrogate pdf to accelerate rather than replace sampling. In \cite{dhamala2018quantifying}, a two-stage model is introduced where a GP surrogate of exact posterior pdf is built in the first stage and is used to improve the acceptance rate of candidate samples in MCMC sampling in the second stage. In \cite{dunbar2020calibration}, a three-stage model is presented for uncertain quantification of a complex climate model parameters where model calibration using Kalman inversion is done in the first stage, building GP surrogate to emulate parameter-to-data map is done in the second stage, and MCMC sampling of the posterior pdf of the climate model parameters is done in the final stage. While these hybrid approaches improve the accuracy of sampling, the reliance on sampling the original pdf limits the extent to which the computation can be reduced. 
\subsection{Parameter Estimation using Active Learning}

Popular active learning algorithms such as efficient global optimization \cite{jones1998efficient}, famously known as Bayesian optimization, have been merged with surrogate modeling to estimate complex model parameters. In Bayesian optimization, a GP surrogate is built to approximate the objective function of the optimization, using a small number of sampling to query the expensive objective function where the samples are selected based on an acquisition function. In many areas such as nuclear physics \cite{ekstrom2019bayesian}, material science \cite{ueno2016combo}, and many more \cite{duris2020bayesian, khosravi2019controller, vargas2019bayesian}, Bayesian optimization is applied to estimate complex model parameters. However, all these techniques are focused on deterministic optimization to find a single optimal parameter value that best fits the simulation output to measurement data without considering the associated uncertainty.

\subsection{Parameter Estimation in Personalized Models}

In the specific area of estimating parameters of patient-specific models, existing works can be classified into deterministic or probabilistic approaches. There are many optimization methods developed in the past few decades. Derivative free methods, such as the Subplex method \cite{wong2015velocity}, Bound Optimization BY Quadratic Approximation (BOBYQA) \cite{wong2012strain}, New Unconstrained Optimization Algorithm (NEWUOA) \cite{sermesant2012patient}, and hybrid particle swarm method \cite{mineroff2019optimization} have been used in estimating cardiac model parameters. Derivative-based variational data assimilation approaches have also been applied to estimate cardiac conductivities in ventricular tissue \cite{yang2015estimation, barone2020experimental}, and heterogeneous elastic material properties in personalized cardiac mechanic model \cite{balaban2018vivo}. Due to the computational expense associated with the model simulation during optimization, model reduction techniques such as Proper Generalized Decomposition (PGD) have been used to accelerate the estimation of cardiac conductivities in personalized cardiac electrical dynamics \cite{barone2020efficient}. These methods overall are focused on finding a single value of cardiac model parameters that best fit the available data, lacking any uncertainty measure associated with the parameters. 

On the other hand, limited progress has been made in the probabilistic estimation of personalized model parameters where the uncertainty measure can be derived from their posterior pdf. To reduce the extensive computation associated with standard MCMC sampling, various approaches of reduced modeling have been used to reduce the cost of forward simulation and thereby accelerate the inverse estimation \cite{lassila2013reduced}. Recent research reports building surrogate models using methods like kriging \cite{schiavazzi2016uncertainty} and polynomial chaos \cite{konukoglu2011efficient} to estimate cardiac model parameters. In \cite{paun2019mcmc}, GP emulation is used to speed up the MCMC process in the area of cardiovascular fluid dynamics. Probabilistic surrogate modeling through GP using Bayesian history matching is applied in \cite{longobardi2020predicting} for inference of cardiac contraction mechanics. In \cite{neumann2014robust}, polynomial chaos method is used to build the surrogate model for fast sampling to estimate parameters of an electromechanical model of the heart. However, with the limited accuracy in the approximated posterior pdf, directly sampling the surrogate results in improved efficacy but reduced accuracy. In \cite{dhamala2018quantifying}, GP surrogate model of the posterior pdf of cardiac model parameters is built to accelerate MCMC sampling of the original posterior pdf. While this strategy avoids the loss of accuracy from sampling the surrogate pdf, it achieves a limited gain of efficiency due to the reliance on MCMC sampling of the original pdf.

\subsection{Estimating High-Dimensional Parameters}

High dimensionality is a bottleneck in estimating parameters, especially in cardiac physiology. Researchers mostly try to explain useful functions through dimension reduction of the original HD parameters. For example, in \cite{malatos2016advances}, it is shown that a lower-dimensional model can be useful in explaining blood flow. In \cite{caruel2014dimensional}, to explain cardiac function, low-dimensional muscle samples or myocytes as model parameters are estimated from high-dimensional ones. Estimating local myocardial infarct uncertainties through reducing the dimension of deformation patterns is introduced in \cite{duchateau2016infarct}. In \cite{giffard2018transfer}, offline learning from electrocardiographic imaging (ECGI) is achieved through dimension reduction of the myocardial shape. As most of the parameters stay on manifold rather than Euclidean space, in \cite{nakarmi2017kernel}, a kernel-based framework using low dimensional manifold models to reconstruct cardiac dynamic MR images is proposed. In \cite{le2016mri}, to reduce dimension, homogeneous tissue excitability (in the form of a model parameter) is represented by a single global model parameter. In \cite{wong2015velocity}, the cardiac mesh is pre-divided into 3-26 segments, each represented by a uniform parameter value. As the number of segments increases, the estimation becomes more challenging and increasingly reliant on initialization. Alternatively, a multi-scale hierarchy of the cardiac mesh is defined for a coarse-to-fine optimization, which allowed spatially-adaptive resolution that was higher in certain regions than the other \cite{dhamala2016spatially, chinchapatnam2008model}. However, the representation ability of the final partition is limited by the inflexibility of the multi-scale hierarchy: homogeneous regions distributed across different scales can not be grouped into the same partition, while the resolution of heterogeneous regions can be limited by the level of scale the optimization can reach \cite{dhamala2017spatially}. In addition, because these methods involve a cascade of optimization along the hierarchy of the cardiac mesh, they are computationally expensive. 

In our recent work, we present an approach that replaces the explicit anatomy-based reduction of the parameter space with an implicit low-dimensional (LD) manifold that represents the generative code for HD spatially-varying tissue excitability \cite{dhamala2018high}. This is achieved by embedding within the optimization a generative model, in the form of a variational autoencoder (VAE) trained from a large set of spatially varying tissue excitability. In our previous work, we demonstrated the efficacy of this approach for deterministic optimization of spatially-varying tissue excitability in cardiac electrophysiological models \cite{dhamala2018high}. In this paper, we leverage this strategy to enable probabilistic estimation of high-dimensional model parameters.

\section{Background}
\label{sec:bck}

\subsection{Bi-ventricular Electrophysiology Model}

There are many computational models with varying levels of biophysical details \cite{clayton2011models, mitchell2003two, aliev1996simple}. Among these, phenomenological models like the Aliev Panfilov (AP) model \cite{aliev1996simple} is capable of reproducing the key macroscopic process of cardiac excitation with a small number of model parameters. To test the feasibility of the presented method, we utilize the two-variable AP model given below:
\begin{equation}
\begin{aligned}
\frac{\partial u}{\partial t}  &= \nabla(\mathbf{D} \nabla u) - cu(u-\theta)(u-1) - uv,\\
\frac{\partial v}{\partial t}  &= \varepsilon(u,v)(-v - cu(u - \theta - 1)).
	\label{eq:AP}
\end{aligned}
\end{equation}
Here, $u \in [0,1]$ is the transmembrane potential and $v$ is the recovery current. The parameter $\varepsilon = e_0 + (\mu_1v)/(u + \mu_2)$ controls the coupling between $u$ and $v$, and $c$ controls the re-polarization. $\mathbf{D}$ is diffusion tensor which controls the spatial propagation of $u$. $\theta$ is tissue excitability parameter that controls the temporal dynamics of $u$ and $v$. Based on previous sensitivity analysis \cite{dhamala2017spatially}, in this study, we focus on estimating parameter $\theta$ of the AP model \eqref{eq:AP}, while fixing the values for the rest of the model parameters based on the literature \cite{aliev1996simple}: $c=8$, $e_0=0.002$, $\mu_1=0.2$, and $\mu_2=0.3$. We solve the AP model~(\ref{eq:AP}) on the discrete 3D myocardium using the meshfree method described in \cite{wang2009physiological}. Then we obtain a 3D electrophysiological model of the heart that describes the spatio-temporal propagation of 3D transmembrane potential $\mathbf{u}(t,\pmb{\theta})$. Note that, compared to existing works where the model parameter to be estimated is often assumed to be global or low-dimensional based on a pre-defined anatomical division of the heart, we consider the estimation of a high-dimensional parameter $\pmb{\theta}$ at the resolution of the cardiac mesh.

In this study, we demonstrate the presented framework using body surface electrocardiogram (ECG) which are generated by spatio-temporal cardiac action potential following the quasi-static approximation of the electromagnetic theory \cite{plonsey2001bioelectric}. In \cite{wang2009physiological}, this relationship is modeled by solving a Poisson's equation within the heart and Laplace's equation external to the heart on a discrete mesh of the heart and the torso, which gives a linear model:
\begin{equation}
\mathbf{Y}_b(t) = \mathbf{H}_b\mathbf{u}(t,\pmb{\theta})
\label{eq:lin_mod}
\end{equation}
Where $\mathbf{Y}_b(t)$ represents ECG data, $\mathbf{u}(t,\pmb{\theta})$ represents transmembrane potential, $\mathbf{H}_b$ is the transfer matrix unique to patient-specific heart and torso geometry, and $\boldsymbol{\theta}$ is the vector of tissue excitability to be estimated.

\section{Methodology}
\label{sec:method}

The electrophysiological system as defined in Section \ref{sec:bck} defines a stochastic relationship between measurement data $\mathbf{Y}$ and model parameter $\boldsymbol{\theta}$ as:
\begin{equation}
\mathbf{Y}=M(\boldsymbol{\theta}) + \boldsymbol{\varepsilon}
\label{eq:stocastic}
\end{equation}
where $M$ is a composite of the whole-heart electrophysiological model and measurement model reviewed in Section \ref{sec:bck}. $\boldsymbol{\varepsilon}$ is the noise term that accounts for measurement errors and modeling errors other than that arising from the value of the parameter $\boldsymbol{\theta}$. Assuming uncorrelated Gaussian noise $\boldsymbol{\varepsilon} \sim N(\boldsymbol{0},\sigma_e^2\boldsymbol{I})$ , the likelihood can be written as:
\begin{equation}
\pi(\mathbf{Y}\vert \boldsymbol{\theta})\propto exp(-\frac{1}{2\sigma_e^2}||\mathbf{Y} - M(\boldsymbol{\theta})||^2)
\label{eq:lik}
\end{equation}
The unnormalized posterior density of the model parameter $\boldsymbol{\theta}$ has the following form, using Bayes rule:
\begin{equation}
\pi(\boldsymbol{\theta} \vert \mathbf{Y})\propto\pi(\mathbf{Y} \vert \boldsymbol{\theta})\pi(\boldsymbol{\theta})
\label{eq:objfunc}
\end{equation}
where $\pi(\boldsymbol{\theta})$ provides us prior knowledge about the parameters. In this study, a uniform distribution bounded within [0, 0.5] is used where the bound is informed by the physiological values of parameter $\boldsymbol{\theta}$. In this general setup, our goal is to estimate the pdf function in \eqref{eq:objfunc}, which has an expensive likelihood function and a HD parameter $\boldsymbol{\theta}$. Naive MCMC sampling of \eqref{eq:objfunc} would render intensive, if not infeasible, computation. Here, we cast the problem of estimating the function of $\pi(\boldsymbol{\theta} \vert \mathbf{Y})$ into a Bayesian active learning problem: we aim to learn a Gaussian process (GP) approximation of the function $\pi(\boldsymbol{\theta} \vert \mathbf{Y})$ from training samples of $\{\boldsymbol{\theta}^{(i)}, \pi(\boldsymbol{\theta^{(i)}} \vert \mathbf{Y})\}_{i=1}^l$; because the evaluation of $\pi(\boldsymbol{\theta^{(i)}} \vert \mathbf{Y})$ involves expensive computation, \textit{i.e.}, an expensive labeling process, we intelligently select a small number of training points $\boldsymbol{\theta^{(i)}}$ on which to query the label of $\pi(\boldsymbol{\theta^{(i)}}\vert \mathbf{Y})$. To achieve this, we bring two innovations to existing Bayesian active learning methods. First, leveraging our previous work \cite{dhamala2017spatially}, we integrate generative modeling of high-dimensional $\boldsymbol{\theta}$ into Bayesian active learning to embed the process of active search of training samples into a low-dimensional manifold. Second, we introduce new acquisition functions for selecting training points $\boldsymbol{\theta^{(i)}}$, such that it focus on the shape of the posterior pdf of interest.

\subsection{Enabling High-Dimensional Bayesian Active Learning via Generative Modeling}
\label{sec:vae_structure}

To obtain a generative model of $\boldsymbol{\theta} = g(\mathbf{z})$, we use VAE that consists of two modules: a probabilistic deep encoder network with network parameters $\pmb{\alpha}$ that approximates the intractable true posterior density $p(\mathbf{z}|\pmb{\theta})$ as $q_{\pmb{\alpha}}(\mathbf{z}|\pmb{\theta})$, and a probabilistic deep decoder network with network parameters $\pmb{\beta}$ that reconstructs $\pmb{\theta}$ given $\mathbf{z}$ with the likelihood $p_{\pmb{\beta}}(\pmb{\theta}|\mathbf{z})$. Given a training data set ${\Theta} = \{\pmb{\theta}^{(i)}\}_{i=1}^N$ that consists of $N$ different spatial distributions of the tissue excitability $\pmb{\theta}$, VAE training involves optimizing the variational lower bound on the marginal likelihood of each training data $\pmb{\theta}^{(i)}$ with respect to network parameters $\pmb{\alpha}$ and $\pmb{\beta}$:
\begin{equation}
\mathcal{L}(\pmb{\alpha};\pmb{\beta}) = -D_{\mathrm{KL}} (q_{\pmb{\alpha}}(\mathbf{z}|\pmb{\theta}^{(i)}) || p(\mathbf{z})) + E_{q_{\alpha}(\mathbf{z}|\pmb{\theta}^{(i)})} [\mathrm{log} p_{\pmb{\beta}}(\pmb{\theta}^{(i)}|\mathbf{z})].
\label{eq:loss}
\end{equation}
We assume the prior $p(\mathbf{z})\sim\mathcal{N}(0,1)$ to be a standard Gaussian density. The optimization of \eqref{eq:loss} with respect to $\boldsymbol{\alpha}$ and $\boldsymbol{\beta}$ is achieved with stochastic gradient descent with re-parameterization trick \cite{kingma2013auto}. After the VAE is trained, the decoder as a generative model can be incorporated into \eqref{eq:objfunc} to obtain:
\begin{equation}
\pi(\mathbf{z}\vert \mathbf{Y})\propto [exp(-\frac{1}{2\sigma_e^2}||\mathbf{Y} - M\big(\mathrm{E}[p_{\pmb{\beta}}(\pmb{\theta}|\mathbf{z})]\big)||^2)] [exp(-\frac{1}{2}||\mathbf{z}||^2)]
\label{eq:postz}
\end{equation}
where $\boldsymbol{\theta}$ is now approximated by the expectation of the generative model $p_{\pmb{\beta}}(\pmb{\theta}|\mathbf{z})$ and the prior of $\mathbf{z}$ is assumed to be Gaussian: $\pi(\mathbf{z}) \sim\mathcal{N}(0,1)$. In another word, the use of $p_{\pmb{\beta}}(\pmb{\theta}|\mathbf{z})$ allows us to now perform Bayesian active learning over the low-dimensional latent space $\mathbf{z}$.

\subsection{Bayesian Active Learning with Posterior-Focused Acquisition Functions}

We aim to learn a GP approximation of the log posterior because, compared to the posterior pdf in \eqref{eq:postz}, it has longer scales and lower dynamic range. In another word, we build a GP to approximate:
\begin{equation}
\mbox{GP}(\mathbf{z}) \sim -\frac{1}{2}(\frac{||\mathbf{Y} - M\big(\mathrm{E}[p_{\pmb{\beta}}(\pmb{\theta}|\mathbf{z})]\big)||^2}{\sigma_e^2} + ||\mathbf{z}||^2)
\label{eq:logpost}
\end{equation}
Bayesian active learning with GP consists of an iterative process. In each iteration, we 1) first select new training samples via the optimization of an acquisition function, and 2) then obtain the posterior distribution of the GP from the prior distribution using newly obtained training samples. For the prior of the GP at the first iteration, we adopt the commonly used zero-mean function due to lack of prior knowledge, and the anisotropic \textquotedblleft Mat\'{e}rn 5/2" covariance function \cite{rasmussen2003gaussian}:
\begin{equation}
\label{eq:kernel}
\mathbf{k}(\mathbf{z}_i,\mathbf{z}_j) = \alpha^2 \textrm{exp} \big(-\sqrt{5}d(\mathbf{z}_i,\mathbf{z}_j)\big) \big(1+\sqrt{5}d(\mathbf{z}_i,\mathbf{z}_j)+5/3d^2(\mathbf{z}_i,\mathbf{z}_j)\big)
\end{equation}
where $d^2(\mathbf{z}_i,\mathbf{z}_j) = (\mathbf{z}_i-\mathbf{z}_j)^T\mathbf{\Lambda}(\mathbf{z}_i-\mathbf{z}_j)$, $\mathbf{\Lambda}$ is a diagonal matrix in which each diagonal element represents the inverse of the squared characteristics length scale along each dimensions of $\mathbf{z}$, and $\alpha^2$ is the function amplitude.

\subsubsection{Acquisition Function Design}

A crucial part of Bayesian active learning is to guide the algorithm about where to sample next, achieved by designing an acquisition function that balances between exploiting what is already learned about the target function of interest and exploring the unknown region of the input space. Existing GP-based Bayesian active learning is typically used for finding the optimum of a target function, using the mean and variance function of the GP approximations of the target function to exploit high-mean regions while exploring high-variance regions. In learning to approximate the posterior pdf function as defined in \eqref{eq:postz}, our goal differs from standard approaches in two ways. First, while we choose to build the GP approximation of the log posterior, we are interested in the accuracy of the posterior pdf function itself as our target function. Second, we are interested in the shape of the posterior pdf, rather than any single optimum value. These motivate the design of new acquisition functions as follows.

First, based on \eqref{eq:postz} and \eqref{eq:logpost}, our posterior pdf of interest is an exponential factor away from the function being approximated by the GP. Since $\mbox{GP}(\mathbf{z})$ at every $\mathbf{z}$ follows a Gaussian distribution, $\mbox{exp(GP}(\mathbf{z}))$ follows log-normal distribution at every $\mathbf{z}$. In another word, the function of $\mbox{exp(GP}(\mathbf{z}))$ follows a log-normal process. To focus on the accuracy of approximating the posterior pdf function, rather than using the mean and variance of the GP to guide acquisition as in regular Bayesian active learning, we will use the mean and variance of the log-normal process $\mbox{exp(GP}(\mathbf{z}))$ to guide acquisition.

Second, to focus more on learning the shape rather than optimum (\textit{i.e.}, mode) of the posterior pdf, we emphasize more on reducing the uncertainty of the learned $\mbox{exp(GP}(\mathbf{z}))$ (\textit{i.e.}, exploration) than exploiting around its mode. Two natural candidates for measuring the uncertainty in the approximated $\mbox{exp(GP}(\mathbf{z}))$ include: 1) variance of $\mbox{exp(GP}(\mathbf{z}))$, and 2) entropy of $\mbox{exp(GP}(\mathbf{z}))$ at any given $\mathbf{z}$:
\begin{equation}
Entropy(\mathbf{z})=\mu(\mathbf{z}) +\frac{1}{2} + ln(\sqrt{2\pi}\sigma(\mathbf{z}))
\label{eq:entropy}
\end{equation}
\begin{equation}
Variance(\mathbf{z})=[\exp{\sigma^2(\mathbf{z})-1}][\exp{2\mu(\mathbf{z})+\sigma^2(\mathbf{z})}]
\label{eq:variance}
\end{equation}
At the $i$-th iteration of active learning, we select a single point of $\mathbf{z}^{(i)}$ that maximizes \eqref{eq:entropy} or \eqref{eq:variance} to update the GP.

\subsubsection{Updating GP with New Training Samples}
\label{sec:update_gp}

Once a new sample point $\mathbf{z}^{(i)}$ is selected, the value of the log posterior in \eqref{eq:logpost} is evaluated at $\mathbf{z}^{(i)}$ as $\pmb{\mathcal{L}}^{(i)}$ which includes the execution of the trained VAE decoder, the bi-ventricular electrophysiological model, and the measurement model as described in Section \ref{sec:bck}. The new input-output pair is used to update the posterior belief of the GP. Following \cite{williams2006gaussian}, the predictive mean and variance of the updated GP can be evaluated at any $\mathbf{z}$:
\begin{equation}
\mu(\mathbf{z}^*) = \mathbf{k}^T\mathbf{K}^{-1}\pmb{\mathcal{L}}^{(1:i)},
\sigma^2(\mathbf{z}^*) = \mathbf{k}(\mathbf{z}^*, \mathbf{z}^*) - \mathbf{k}^T\mathbf{K}^{-1}\mathbf{k}
\label{eq:gp_mean_var}
\end{equation}
where $\mathbf{k}$ is the kernel function. We update the kernel hyperparameters, including the length-scale and noise variance mentioned in \eqref{eq:kernel}, every time we add a new training point by maximizing the log of the marginal likelihood. 

Overall, the active learning process involves two steps: 1), Adding new training points by maximizing the acquisition function, and 2), Updating the GP posterior mean and variance function. This iterative process continues until the Kullback–Leibler (KL) divergence between the most updated predictive mean pdf function and the average of the last five predictive mean pdf functions of $\mbox{exp(GP}(\mathbf{z}))$ does not exceed a predefined threshold. The length-scale and noise variance of kernel function are optimized every time by maximizing log of the marginal likelihood. 

\section{Experiments and Results}
\label{sec:experiments}

\subsection{Generative Modeling of Spatially-Varying Tissue Excitability}
\label{subsec:VAE}

Tissue excitability of whole-heart from real data is not readily available. Cardiac images such as contrast-enhanced MRI may provide a surrogate for delineating different levels of myocardial injury, yet they are expensive to obtain at a large quantity. In this study, we utilized synthetic data sets $\pmb{\Theta}=\big\{\pmb{\theta}^{(i)}\big\}_{i=1}^{N}$ to train the VAE. Specifically, we generated a large data set of heterogeneous myocardial injury by random region growing. Starting with one injured node, one out of the five nearest neighbors of the present set of injured nodes was randomly added as an injured node. This was repeated until an injury of desired size was attained. We considered binary tissue types in the training data, in which the value of tissue excitability $\boldsymbol{\theta}$ was set to be $0.5$ or $0.15$ for injured or healthy nodes, respectively, along with a random noise drawn from a uniform distribution $[0,0.001]$.

\begin{figure}[t!]
\begin{center}
   \includegraphics[width=1\textwidth]{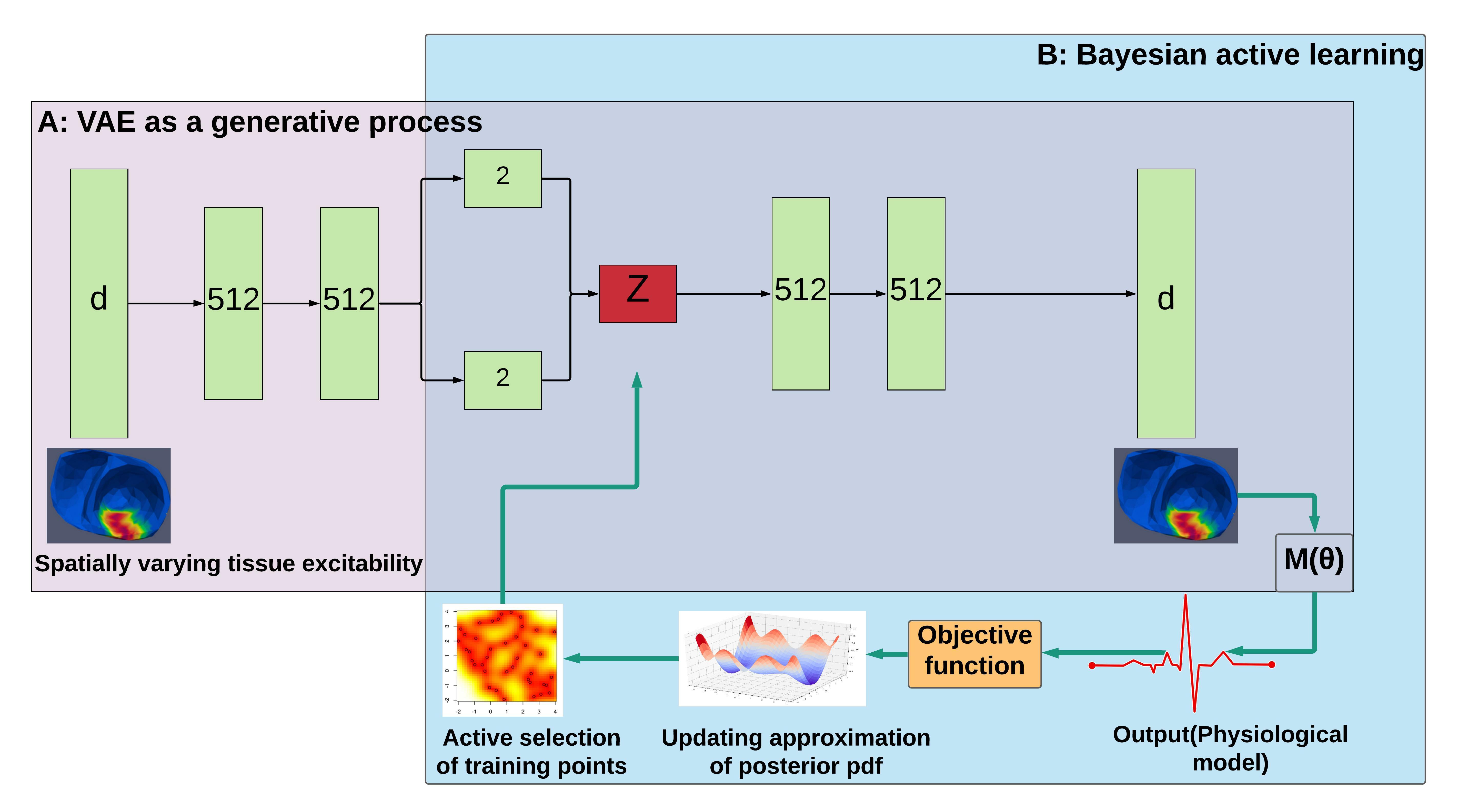}
   \end{center}
    \caption{Workflow of the presented method. A: A generative model of high-dimensional spatially-varying tissue excitability of the 3D heart is trained offline; B: The resulting generative model is embedded into Bayesian active learning to approximate the posterior pdf of model parameters using a small number of intelligently selected training points guided by the acquisition function.}
    \label{fig:work_flow}
\end{figure}

The VAE architecture used in the following experiments is shown in Fig.~\ref{fig:work_flow}. Each of the encoder and decoder network consisted of three fully connected layers with softplus activation, two layers of 512 hidden units, and a pair of two-dimensional units for the mean and log-variance of the latent code $\mathbf{z}$. We trained the VAE with the Adam optimizer with an initial learning rate of 0.001 \cite{kingma2013auto}.

\begin{figure}[t!]
\begin{center}
   \includegraphics[width=1\textwidth]{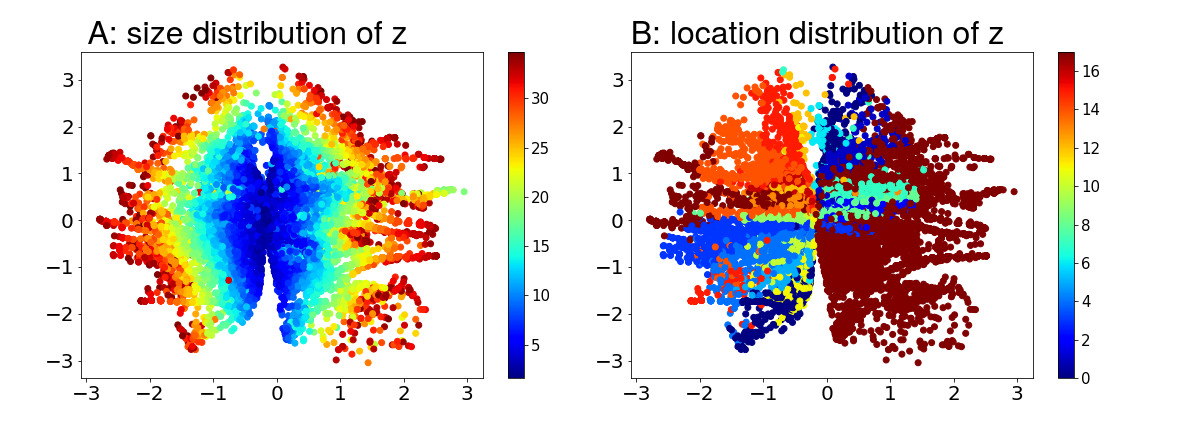}
   \end{center}
    \caption{{Distribution of low-dimensional latent codes of the training data, color coded by A: size of the abnormal tissue (the colors represent the percentage size of abnormal tissue); B: location of the abnormal tissue (the colors represent the 17  American Heart Association (AHA) segments of left ventricle).}}
    \label{fig:train_vae}
\end{figure}

Fig.~\ref{fig:train_vae}A-B shows the scattered plots of the two-dimensional latent codes $\mathbf{z}$ encoded by the VAE on the training data, color-coded by the size and location of the abnormal tissue. It appears that the latent code accounted for the size of the abnormal tissue along the radial direction (A), while clustering by the location of the abnormal tissue as well (B). This shows the ability of the generative model in capturing meaningful semantic information in the high-dimensional data in an unsupervised manner.

\subsection{Synthetic Data Experiments}
\label{subsec:syn_experiments}

Synthetic experiments were carried out on three computed tomography (CT)-derived human heart-torso models. For ground truth of the tissue excitability, we divided the left ventricle (LV) into 17 segments based on the standard recommended by the American Heart Association (AHA). The region of abnormal tissue was then set as various combinations of these 17 LV segments. The value of $\boldsymbol{\theta}$ in the abnormal region was set to $0.40$, $0.45$, or $0.50$ to have different severity levels, and its value in the healthy region was set to $0.15$. A random noise drawn from a uniform distribution [0, 0.001] was added. Note that the tissue excitability in this test set is different from those in the training set, as described in Section \ref{subsec:VAE}, in two aspects: 1) parameter values within the abnormal region and 2) shape and size of the abnormal region.

For each tissue excitability to be tested, body-surface measurements were simulated using the models described in Section \ref{sec:bck}. 20dB noise was then added to the measurement data for posterior estimation of parameter $\boldsymbol{\theta}$. To test the ability of the trained VAE model to be applied to hearts different from that used in training, for experiments on heart $\sharp 1$ and $\sharp 2$, the VAE was trained on heart $\sharp 3$; for experiments on heart $\sharp 3$, the VAE was trained on heart $\sharp 1$. The convergence criteria for each estimation followed that as defined in Section \ref{sec:update_gp}.

\subsubsection{Accuracy and Efficiency in Estimating Posterior PDF Function}

We first evaluated the accuracy and efficiency of the presented method against 1) directly sampling GP approximation of the posterior pdf based on regular Bayesian active learning, and 2) surrogate-accelerated two-stage MCMC sampling as presented in our previous work \cite{dhamala2017quantifying}, all against the baseline of directly sampling the exact posterior pdf using the standard MCMC.  We considered 15 synthetic cases in total. All MCMC sampling were run on two parallel MCMC chains of length 10,000 with a common Gaussian proposal distribution with two different initial points. The variance of the Gaussian proposal distribution was tuned by rapidly sampling the GP surrogate pdf until obtaining an acceptance rate of 0.22, which is documented to enable good mixing and faster convergence in higher dimensional problems (Andrieu et al., 2003; Gilks et al., 1995). After discarding 20 percent initial burn-in samples and selecting alternate samples to avoid auto-correlation in each chain, the samples from two chains were combined. The convergence of all the MCMC chains was tested using trace plots, Geweke statistics, and Gelman-Rubin statistics (Andrieu et al., 2003; Gilks et al.,1995). 
 
\begin{table}
\caption {Table 1: Absolute errors in the estimated mean, mode, and standard deviation of the estimated posterior pdf and its KL divergence against directly sampling the exact posterior pdf: the presented method vs. sampling the surrogate from regular Bayesian active learning (regular BAL) vs Two-stage MCMC} \label{tab:title} 
\begin{tabular}{ |c|c|c|c|c| } 
\hline
Method & Mean & Mode & Standard deviation & KL divergence\\
\hline
Two-stage MCMC & 0.04 $\pm$ 0.003 & 0.02 $\pm$ 0.001 & 0.08 $\pm$ 0.002 & 0.3 $\pm$ 0.1 \\
\hline
Regular BAL & 0.4 $\pm$ 0.1 & 0.03 $\pm$ 0.004 & 1.1 $\pm$ 0.2 & 0.9 $\pm$ 0.25 \\
\hline
Presented method & 0.1 $\pm$ 0.02 & 0.02 $\pm$ 0.002 & 0.12 $\pm$ 0.03 & 0.6 $\pm$ 0.2  \\
\hline
\end{tabular}
\end{table}

\begin{figure}[t!]
\begin{center}
   \includegraphics[width=1\textwidth]{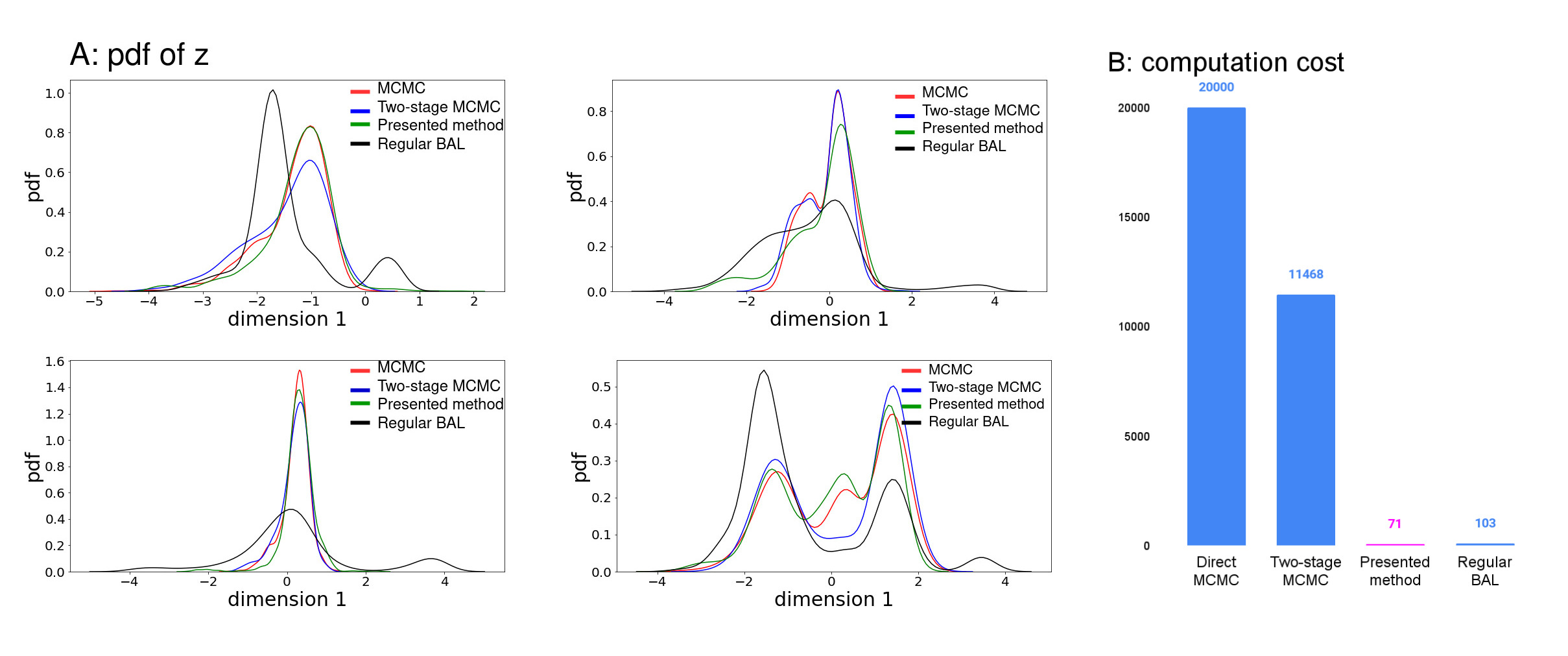}
   \end{center}
\caption{A: Comparison of estimated posterior pdf from different methods; B: Comparison of computation cost from different methods.}
\label{fig:curves}
\end{figure}

The accuracy of estimated pdf in $\mathbf{z}$ space was evaluated through comparing the mean, mode, and standard deviation from the kernel density estimation of samples selected from our method and with other existing methods. Let $s_M$ be the estimated mean, mode, or standard deviation of the posterior pdf of $\mathbf{z}$ using direct MCMC sampling and $s_o$ be the corresponding statistics estimated from the three methods presented in Table 1. We used the mean and standard deviation of $|s_M - s_o|$ calculated from 15 synthetic cases to evaluate the accuracy of all the comparison methods in estimating the mean, mode, and standard deviation of the posterior pdf in comparison to the direct MCMC sampling. The last column of Table 1 also showed the KL divergence between the estimated pdf from different methods with that from exact MCMC, obtained by sampling as described in \cite{hershey2007approximating}. As shown, the accuracy of the estimated posterior pdf was significantly higher than that obtained by regular Bayesian active learning (paired t-test on estimated parameters from 15 cases, p $<$ 0.001). While its accuracy was still lower than the surrogate-accelerated two-stage MCMC, it used only 0.6\% computation (in terms of the number of model simulations needed) of the two-stage MCMC method. As detailed in Fig.~\ref{fig:curves}B, while the two-stage MCMC achieved $\sim 40\%$ reduction of the number of model simulations needed compared to the direct sampling of the exact posterior pdf, the presented method reached a $\sim 99.65\%$ reduction of computation. Fig.~\ref{fig:curves}A gives examples of the posterior pdfs estimated from different methods in comparison to that obtained from direct sampling.

As shown, the presented method (green curve) closely reproduced the true posterior pdf (red curve) obtained from direct MCMC, while the function learned by the standard Bayesian active learning (black curve) fell short in as closely reproducing the posterior pdf.

\subsubsection{Accuracy and Uncertainty in the Estimated Tissue Excitability}

From the estimated posterior pdf of $\pi(\mathbf{z}|\mathbf{Y})$ over the latent LD manifold, we obtained the posterior pdf of $\pi(\boldsymbol{\theta}|\mathbf{Y})$ over the spatial space of the heart. We estimated the mean, mode, and standard deviation in HD space through inserting MCMC samples of $\mathbf{z}$ taken from posterior $\pi(\mathbf{z}|\mathbf{Y})$ to the expectation network of the trained VAE decoder.

\begin{figure}[t!]
\begin{center}
   \includegraphics[width=1\textwidth]{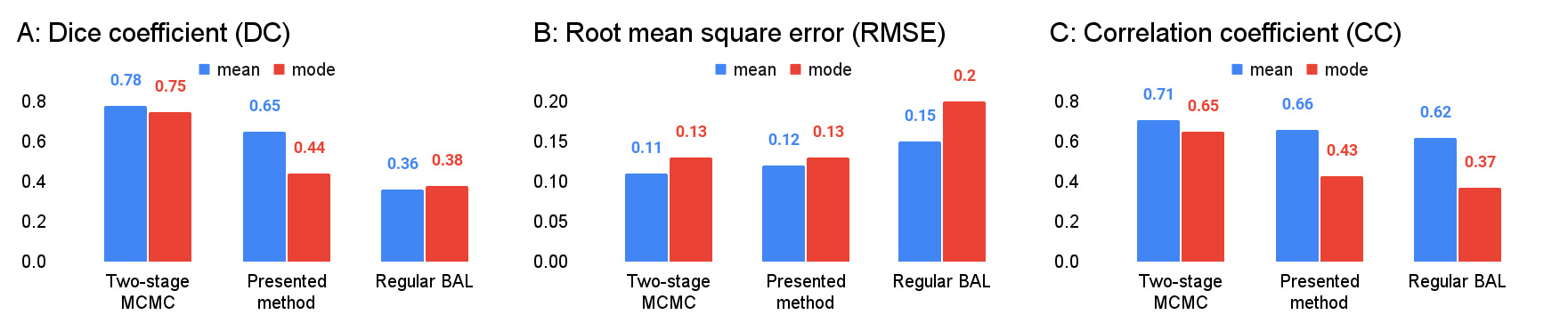}
   \end{center}
\caption{Comparison of A: DC, B: RMSE, and C: CC between estimated mean (blue) or mode (red) tissue excitability in comparison to the ground truth.}
\label{fig:dc_rmse_cc}
\end{figure}

For accuracy of the estimated tissue excitability, we considered the mean and mode from the estimated posterior pdf of $\pi(\boldsymbol{\theta}|\mathbf{Y})$, and evaluated against the ground truth tissue excitability using three metrics: Dice coefficient (DC), root mean square error (RMSE), and correlation coefficient (CC). As shown in Fig. \ref{fig:dc_rmse_cc}, for DC, the mean and mode from the presented method were more accurate than those obtained by regular Bayesian active learning (paired t-test, $p<0.001$ for mean and $p<0.05$ for mode), but less accurate than those obtained from the two-stage MCMC (paired t-test, $p<0.10$ for mean and $p<0.001$ for mode). For RMSE, similarly, mean and mode both were more accurate from regular active learning method (paired t-test, $p<0.005$ for mean and $p<0.05$ for mode). In comparison with the two-stage MCMC, there was no difference for mean and mode with the presented method (paired t-test, insignificant at 20\% level of significance). For CC, our presented method showed similar accuracy with the two-stage MCMC and regular active learning method for mean estimation. But for CC from mode estimation, our method showed higher accuracy than the regular method (paired t-test, $p<0.01$) but less accuracy than the two-stage MCMC (paired t-test, $p<0.05$).

\begin{figure}[t!]
\begin{center}
   \includegraphics[width=1\textwidth]{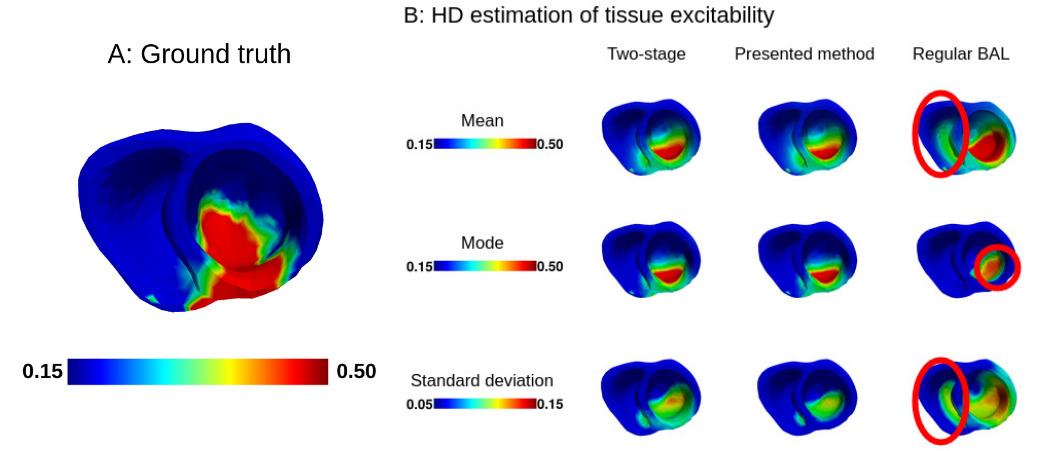}
   \end{center}
\caption{A: The ground truth of tissue excitability; B: Mean, mode, and standard deviation of tissue excitability estimated from presented method.}
\label{fig:3d_heart}
\end{figure}

Fig. \ref{fig:3d_heart}A provides a visual example of the estimated spatially-varying tissue property on the heart, corresponding to the low-dimensional posterior pdf shown in the left column of Fig. \ref{fig:curves}A. First, as shown, the estimated mean provided by the presented method corrected a false positive in the solution from regular Bayesian active learning (row one). The high uncertainty in this region from the regular Bayesian active learning was also corrected by the presented method (row three). Second, as noted in the left column of Fig. \ref{fig:curves}A, the underlying low-dimensional posterior pdf is uni-modal, where both the presented method and two-stage MCMC correctly recovered the mode in comparison to regular Bayesian active learning. Similarly, the resulting mode in the high-dimensional space of the tissue property, was correctly located in position in the presented method whereas the mode of regular Bayesian active learning shifted in accordance with low dimensional shift. This shows a correct one-to-one mapping of LD to HD generative process. Finally, as noted earlier, while the two-stage MCMC, in general, delivered higher accuracy, this performance gain was achieved with over 167-fold increase in computation.

\subsubsection{Exploration vs Exploitation Using Log-Normal Processed Based Acquisition Functions} 

\begin{figure}[t!]
\begin{center}
   \includegraphics[width=1\textwidth]{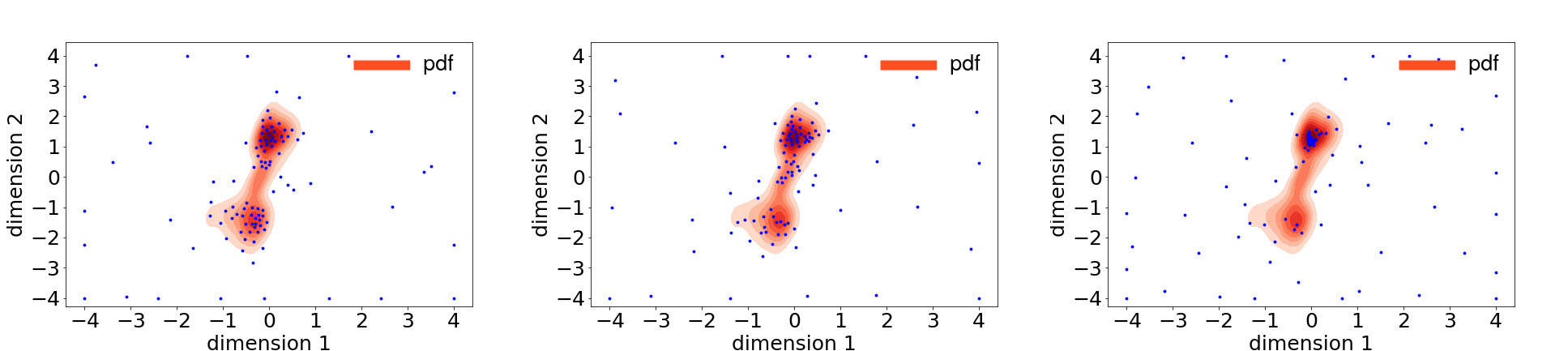}
   \end{center}
\caption{Illustrations of training points (blue dots) selected using variance based on the log-normal process (left), entropy based on the log-normal process (middle), and upper confidence bound (UCB) based on the Gaussian process (right).}
\label{af_plots}
\end{figure}

To understand the advantage of the presented lognormal process based acquisition functions, we examined where the active selection of training samples took place in the presented method versus regular Bayesian active learning. Fig. \ref{af_plots} left and middle showed the acquisition of training samples using the variance and entropy of the lognormal process, using respectively 100 and 108 sampling points to meet the convergence criteria. The contour plot inside these figures showed the shape of the true bivariate posterior pdf. In comparison, Fig. \ref{af_plots} right panel showed training samples selected based on the GP using upper confidence bound (UCB). To converge, it took 129 acquisition steps which was higher than those used in the presented method. Comparing left and middle panel, it showed that the regular acquisition, while exploited the mode of the posterior mode, explored without focusing on the posterior support. In comparison, the presented acquisition functions effectively both exploited and explored within the posterior support.

\subsection{Experiments on Post-infarction Hearts with Blinded Simulation Data}
\label{subsec:jhu}
\subsubsection{Experimental Data and Data Processing}

In this section, we increased the difficulty of active posterior estimation by: 
1) considering hearts with realistic tissue excitability extracted from contrast-enhanced MRI (CE-MRI), and 2) simulation data of 3D cardiac electrical activity generated by a high-fidelity biophysics model blinded to the AP model used in the active posterior estimation. In comparison to synthetic data considered in Section \ref{subsec:syn_experiments}, these image-derived tissue excitability had the following characteristics that increased its heterogeneity: the presence of 1) both dense infarct core and gray zone, 2) a single or multiple infarcts with complex spatial distribution and irregular boundaries, and 3) both transmural and non-transmural infarcts.

We considered six post-infarction human hearts. The patient-specific ventricular models along with the detailed 3D infarct architectures were delineated from MRI images as detailed in \cite{arevalo2016arrhythmia}. The training of VAE was performed on one of the hearts described in Section \ref{subsec:VAE}, using synthetically-generated tissue excitability values as described in that section.

\begin{figure}[t!]
	\centering
 		\includegraphics[width=0.95\textwidth]{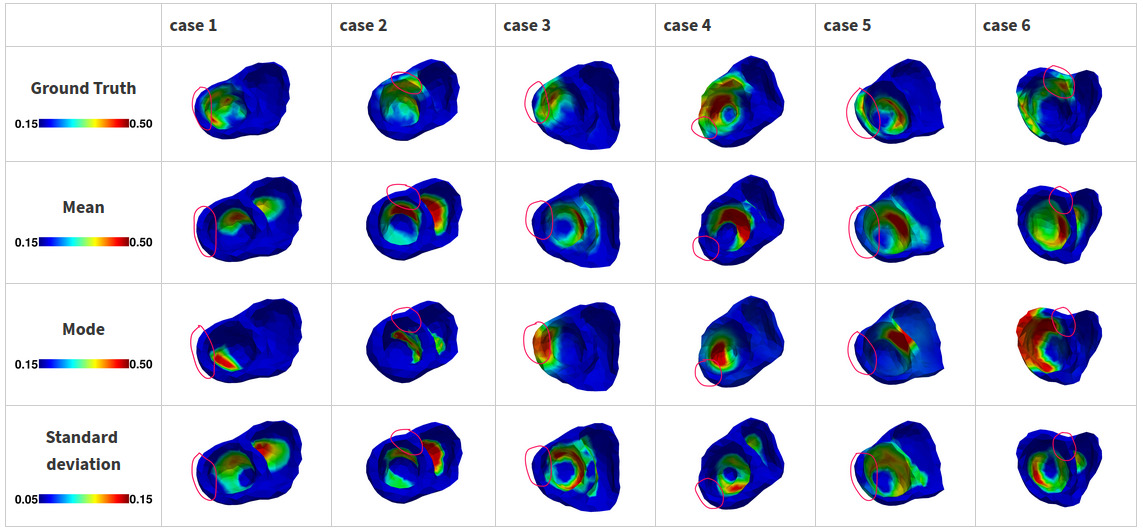}
    \caption{Results of estimated tissue excitability from the presented method in 3D infarcts delineated from \textit{in-vivo} MRI images. Regions with low excitability (high $\pmb{\theta}$ values) correspond to infarct regions (0.5 = infarct core, 0.3-0.5 = gray zone). The red circles highlight non-transmural scars or gray zone.} 
\label{fig:jhu}
\end{figure}

Fig.~\ref{fig:jhu} summarizes the results of estimated tissue excitability on the six post-infarction hearts. Overall, estimated tissue property, especially the estimated mode, was close to the ground truth. One more source of increased difficulty in this set of experiments, in comparison to those in synthetic data, was the presence of non-transmural scar tissue that did not exist in the training data of the VAE. This difficultly in estimating has been previously reported in literature~\cite{dhamala2017spatially}. As shown in Fig.~\ref{fig:jhu} cases 1-3 and 5 (second and third rows), the estimated mean or mode either missed the region of non-transmural abnormal tissue property, or incorrectly estimated it to be transmural (case 3 - mode). The associated uncertainty was not captured in the estimated standard deviation (Fig.~\ref{fig:jhu} (fourth row)) either. Another source of difficulty is the presence of diffused heterogeneous abnormal tissue that was not considered in the VAE training data. For instance, in case 4 and case 6, there was a large patchy grey zone mixed within the dense scars. These regions were reflected in the region of estimated abnormal tissue excitability, however the estimated parameter values were not able to distinguish between the gray zone and dense infarct. In addition to identifiability issues associated with the presented method and the available data, this performance may also arise from the fact that the AP model considered has limited ability in differentiating electrical behavior from gray zone and infarct core \cite{ramirez2020role}.

\subsection{Experiments on \textit{in-vivo} ECG and voltage mapping data}

\begin{figure}[t!]
	\centering
 		\includegraphics[width=0.95\textwidth]{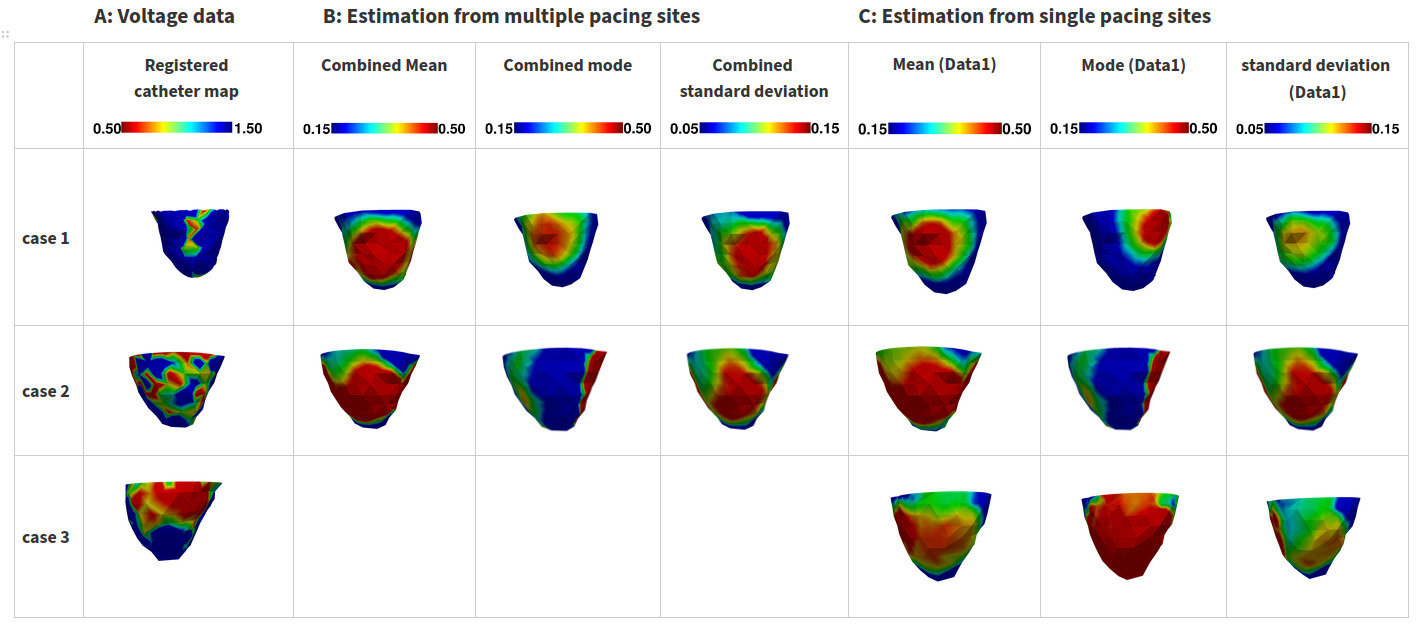}
    \caption{Results of estimated tissue excitability from the presented method in real clinical data. A: Voltage data from catheter map; B: Mean, Mode, and Standard deviation estimated from multiple observations from different pacing sites; C: Mean, Mode, and Standard deviation estimated from a single observation from one pacing site.} 
\label{fig:clinical}
\end{figure}

Finally, we performed active posterior estimation for tissue excitability in real data experiments of three patients who went catheter ablation of ventricular tachycardia due to myocardial infarction \cite{sapp2012inverse}. The patient-specific geometrical models of the heart and torso were constructed from axial CT images detailed in \cite{wang2016noninvasive}. \textit{In-vivo} measurements of 120-lead ECG were collected during pacing from known sites of each heart. The surrogate used for evaluating the estimated tissue excitability was \textit{in-vivo} bipolar voltage data collected by catheter mapping. As illustrated in Fig. \ref{fig:clinical}, based on the voltage data, the myocardium tissue can be divided into three groups: infarct core (red: bipolar voltage $<$ 0.5 mv), infarct border (green: bipolar voltage 0.5-1.5 mv), and healthy (blue: bipolar voltage $>$ 1.5 mv). Among the three patients, we consider 120-lead ECG data collected from a total of six different pacing sites.

\textit{1) Case 1:} 
In this case, we were able to estimate the posterior pdf of tissue excitability by combining ECG data from two different pacing locations. As shown in Fig.~\ref{fig:clinical}A (first row), this subject had a small infarct in the lateral-basal area of left ventricle (LV). The presented method was able to capture the location of this infarct core, although much more smoothed out in comparison to the voltage data as illustrated in Fig.~\ref{fig:clinical}B (first row)). The estimated pdf also exhibited uncertainty higher than the rest of the myocardium in this location. These results were obtained by 129 active acquisitions of simulations with the presented method. 

Interestingly, when estimating the posterior pdf using only data from one pacing location, the mode of the estimated pdf was incorrectly shifted from the actual location of the infarct tissue -- the uncertainty at that location correspondingly became higher compared to that associated with estimation using multiple ECG data (Fig.~\ref{fig:clinical}C (first row)). 

\textit{2) Case 2:} 
In this case, we were able to estimate the posterior pdf of tissue excitability by combining ECG data from three different pacing locations. As illustrated in Fig. \ref{fig:clinical}A (second row), this subject had a highly heterogeneous infarct in the lateral region of the LV. The presented method, using 153 active acquisitions of simulations, was able to recover the correct location of the infarct, with an attempt to recover the heterogeneity in the tissue excitability (Fig. \ref{fig:clinical}B (second row)). The mode solution was also shifted from the target region. The heterogeneity however was not captured in fine detail, likely due to the lack of such heterogeneous data in the VAE training. The associated uncertainty of the solution was accordingly high. When reducing the measurement data to only ECG data from one pacing site, the estimated solution is almost similar when we used three pacing sites.

\textit{3) Case 3:} 
In this case, we only had access to one-paced ECG data for estimating the posterior pdf of tissue excitability. As illustrated in Fig. \ref{fig:clinical}A (third row), this case had a relatively dense scar in inferolateral LV with only one set of measurement data. The presented method was able to locate the infarct using 147 active acquisitions of simulations, with an uncertainty lower than that of the previous two cases (Fig. \ref{fig:clinical}C (third row)). 

\section{Limitations and Future Works}

In this paper, we demonstrated the feasibility of Bayesian active learning for fast approximation of posterior pdf involving heavy simulations. Our key innovation was to modify the acquisition functions in regular Bayesian active learning, such as to focus more on approximating the shape of the posterior pdf of interest rather than finding the mode of the pdf when using regular acquisition functions. Following this idea, in this paper, we demonstrated the feasibility of guiding acquisition with the variance or entropy of the log-normal process being learned. Future work will continue to explore this idea in other acquisition functions, with a goal to modulate the trade-off between exploitation and exploration over the space of $\mathbf{z}$ based on the prior knowledge of its distribution. One possible example is to consider the improvements in the KL divergence between the actual and approximated posterior pdf.

While the parameter $\pmb{\theta}$ was represented in Euclidean space in this study, organ tissue excitability is actually defined over a physical domain in the form of a 3D geometrical mesh. By representing this non-Euclidean data in a Euclidean space, we have ignored the 3D spatial structure of the physical mesh. A future step would be to construct the generative model in non-Euclidean space by considering the geometrical mesh as a graph \cite{dhamala2019bayesian}. We fixed other parameters values in the electrophysiological model in \eqref{eq:AP} to estimate $\pmb{\theta}$; while a better strategy could be varying all the parameters through respective distributions \cite{niederer2020creation}. As a feasibility study, we considered a scalar parameter per cardiac mesh node; this simplifies the problem, although the parameter space was still high dimensional since the parameter values change across space. Future study should consider diffusion tensor $\mathbf{D}$, which requires considering fiber directions that are largely approximated and associated with errors. The lack of real data of organ tissue excitability is the main challenge for training the generative model. A natural next step is to investigate the possibility of using accessible tissue excitability data derived from \textit{in-vivo} and \textit{ex-vivo} optical mapping \cite{kappadan2020high, gizzi2013effects, uzelac2021quantifying}. In this paper, the VAE was trained by synthetic data only that is simplified in shape, transmurality, and heterogeneity. It thus may have a limited ability to generalize to realistic conditions where tissue abnormality is more complex in these aspects. An important direction of future work is to investigate means to improve the training data for the generative model.

While the VAE provides a probabilistic generative model $p_\beta(\pmb{\theta}|\mathbf{z})$, we only adopted the expectation network of this probabilistic model, $\mathrm{E}[p_\beta(\pmb{\theta}|\mathbf{z})]$, as the generative model to achieve the HD-to-LD embedding of the optimization objective. An immediate next step is to investigate the incorporation of the uncertainty in the  generative model into both the active learning of $\pi( \mathbf{z}\vert \mathbf{Y})$ and the estimated pdf $\pi(\boldsymbol{\theta} \vert \mathbf{Y})$. 

Finally, this work focuses on the specific component of tissue excitability estimation within the much bigger pipeline of personalized cardiac modeling. We thus focused on validating the estimated tissue excitability using synthetic as well as \textit{in-vivo} imaging and mapping data. A next step will be to evaluate the personalized model in predictive tasks, such as predicting the risk~\cite{arevalo2016arrhythmia} or the optimal treatment target ~\cite{trayanova2018personalized} for lethal ventricular arrhythmia, and investigate how the uncertainty propagates to simulation outputs and may impact clinical decisions.

\section{Conclusions}

In this paper, we present a novel framework for fast approximation of the posterior pdf of high-dimensional simulation parameters through intelligently selecting training points. This is achieved by casting posterior inference into the setting of Bayesian active learning, integrated with 1) generative modeling to allow active search over high-dimensional parameter space and 2) novel acquisition functions to focus on the shape rather than modes of the posterior pdf. Future work will investigate the design of additional acquisition functions, the incorporation of the uncertainty in the generative model, and the extension of the presented methodology to probabilistic estimation in other complex simulation models.


\end{document}